\pdfoutput=1
\documentclass[11pt]{article}

\usepackage[preprint]{acl}

\usepackage{times}
\usepackage{latexsym}
\usepackage{array}
\usepackage{caption}
\usepackage{graphicx}
\usepackage{svg}
\usepackage{diagbox}
\usepackage{fancyhdr}

\usepackage[T1]{fontenc}
\usepackage[utf8]{inputenc}

\usepackage{microtype}

\usepackage{inconsolata}

\title{Krutrim LLM: A Novel Tokenization Strategy for Multilingual Indic Languages with Petabyte-Scale Data Processing}

\author{Author 1 \and Author n \\
        Address line  \\ Address line}

\author{ 
        \textnormal{Rahul Kumar,
        Shubham Kakde,
        Divyansh Rajput, Daud Ibrahim},\\
        Rishabh Nahata,
        Pidathala Sowjanya,
        Deepak Kumar,
        Gautam Bhargava,
        Chandra Khatri* 
}
\fancypagestyle{firstpage__}{%
  \fancyfoot[L]{\textit{Additional contributors:} Soham Pendurkar, Sindhu Pawar, Arveti Manjunath, Vinayak Dhruv, Aditya Kallappa, Palash Kamble}
}

\fancypagestyle{firstpage}{%
  \fancyfoot[C]{\textit{*Additional contributors:} Soham Pendurkar, Sindhu Pawar, Arveti Manjunath, Vinayak Dhruv, Aditya Kallappa, Palash Kamble, Akshat Patidar, Pranav Taneja, Pranav Raveendran, 
  Bidyapathi Ray, Azhagiri S, Priyanka Nayak}
  \fancyhead{} % Clear all header fields
  
}

\begin{document}

\maketitle

\begin{abstract}
% We present a groundbreaking advancement in natural language processing for multiple Indic languages, achieved through meticulous data acquisition and preprocessing. Our data sources include Common Crawl, Indic books (in PDF format), news outlets, and Wikipedia, ensuring a diverse and rich linguistic representation. For each Indic language, we designed a custom preprocessing pipeline to effectively eliminate redundant and low-quality content. Additionally, we performed deduplication on Common Crawl data to address the redundancy present in 70\% of the crawled web pages. 
% This study focuses on developing high-quality data, optimizing tokenization for our multilingual dataset, and training proprietary 3-billion-parameter and 7-billion-parameter models engineered for superior performance in Indic languages. We introduce a novel multilingual tokenizer training strategy, demonstrating that our custom-trained Indic tokenizer outperforms the state-of-the-art OpenAI Tiktoken tokenizer, achieving a superior token-to-word ratio for Indic languages. 
% Furthermore, we conducted extensive pretraining experiments with 3B and 7B models, exploring various levels of data cleaning, preprocessing, and data mixes. The findings indicate that the loss converges significantly faster and to a much lower value when the data is more refined.

We present a novel approach to data preparation for developing multilingual Indic large language model. Our meticulous data acquisition spans open-source and proprietary sources, including Common Crawl, Indic books, news articles, and Wikipedia, ensuring a diverse and rich linguistic representation. For each Indic language, we design a custom preprocessing pipeline to effectively eliminate redundant and low-quality text content. Additionally, we perform deduplication on Common Crawl data to address the redundancy present in 70\% of the crawled web pages.
This study focuses on developing high-quality data, optimizing tokenization for our multilingual dataset for Indic large language models with 3B and 7B parameters, engineered for superior performance in Indic languages. We introduce a novel multilingual tokenizer training strategy, demonstrating our custom-trained Indic tokenizer outperforms the state-of-the-art OpenAI Tiktoken tokenizer, achieving a superior token-to-word ratio for Indic languages.
% Furthermore, we conducted extensive pretraining experiments with 3B and 7B models, exploring various levels of data cleaning, preprocessing, and data mixes. The findings indicate that the loss converges significantly faster and to a much lower value when the data is more refined.
\end{abstract}  % Rahul
\thispagestyle{firstpage}

\section{Introduction}

In the ever-evolving realm of Natural Language Processing (NLP), the development of Large Language Models (LLMs) have seen a meteoric rise since the inception of transformers. The introduction of LLMs has ushered in a new era in NLP, where the boundaries of what machines could achieve with human language are constantly pushed to astonishing limits. OpenAI's ChatGPT and Google’s BARD has been a pioneering force for redefining the landscape of language modeling performance and has also illuminated the vast societal implications inherent in these technological advancements. Alongside ChatGPT and BARD, various open source and proprietary LLMs have showcased remarkable natural language understanding and generation. 
    There have been public releases by some key players in the realm of LLM including Llama 2 \cite{touvron2023llama}, Mistral \cite{jiang2023mistral}, Falcon, MPT, etc creating more affordable, efficient, and high-performing language models. The chat-models of these pretrained LLMs showcase comparable results on few benchmarks with proprietary LLM chat-models including ChatGPT, BARD and Claude. LLMs such as Llama and Falcon have outlined their data refinement and pre-training steps in their respective technical reports. We draw inspiration from the meticulous analysis of RefinedWeb Dataset by Falcon in our data filtering and deduplication approaches. 
    
Indic culture boasts linguistic diversity, encompassing Indo-Aryan, Dravidian and Munda languages. Indo-Aryan and Dravidian languages constitute 96\% of India's spoken languages. Despite this richness, most open language models lack Indic language support, hindering innovation due to limited high-quality data and complex tokenization challenges \cite{OpenHathi}.
Notable corpora like EMILLE/CIIL, Wikipedia for Indian languages, Samantar Corpus, and AI4Bharat-IndicNLP provide valuable resources \cite{kunchukuttan2020ai4bharat}. EMILLE/CIIL spans 14 languages with 92 million words, Wikipedia for Indian languages is limited, Samantar Corpus offers 49.7 million parallel sentences across 11 languages, and AI4Bharat-IndicNLP Corpus contains 2.7 billion words for 10 languages \cite{kunchukuttan2020ai4bharat}. IndicCorp with 8.8 billion tokens across 11 languages, supplements linguistic resources \cite{kakwani2020indicnlpsuite}. Despite these efforts, it is noteworthy that Hindi being the third most spoken language, does not rank among the top 20 languages in processed CommonCrawl documents, highlighting the scarcity of India-specific data for open large language model training \cite{Penedo2023}.

In this study, we provide a technical report on clean indic dataset preparation and tokenizer training for Indic LLM : India’s own foundational model with indic-rich context. The study highlights a comprehensive analysis of available open source and proprietary datasets and data refinement steps. We also devise a state-of-art indic tokenizer through rigorous experimentations and validated the performance through a pretraining model.
 % Kakde and Divyansh

\section{Related Works}

Large language models (LLMs) owe their remarkable learning capabilities to massive model sizes and extensive training datasets. At present, there are numerous foundational models spanning from open source and proprietary LLMs. Some noteworthy open source foundational LLMs include Llama 2 \cite{touvron2023llama}, Mistral 7B \cite{jiang2023mistral}, Falcon \cite{Penedo2023}, MPT \cite{MosaicML2023Introducing}, Bloom \cite{workshop2022bloom}, etc. whereas proprietary foundational LLMs comprise of GPT4 \cite{achiam2023gpt}, LaMDA \cite{thoppilan2022lamda}, etc. LLaMA 2 developed by Meta AI and Microsoft focuses on multilingual capabilities and is optimized for swift training and inference. MPT-7B by MosaicML and Mistral-7B by Mistral AI are 7-billion-parameter models that have demonstrated efficient open-source training code, promoting transparency and ease of use. These models have showcased superiority over other open-source models in the 7B-20B range. Falcon-40B developed by Technology Innovation Institute (TII) has 40-billion-parameters and is a causal decoder-only model trained on a causal language modeling task. It is trained on a large dataset and has demonstrated superior performance to GPT-3. BLOOM is the world's largest open multilingual language model, with 176-billion-parameters. Generally, proprietary systems are more expensive and offer product solutions that can be tailored to fit very specific business needs. Open source models usually offer more affordable and customizable options but may lack the performance level and specialization of proprietary LLMs. 

    Despite the widespread availability of LLMs for public exploration, the lack of transparency regarding training datasets, especially for state-of-the-art models, hinders research on addressing relevant biases. Furthermore, LLMs are known to generate text lacking sufficient grounding to knowledge sources, thus posing risks of misinformation and hallucination \cite{li2023halueval}. This challenge is exacerbated largely in multilingual learning scenarios where datasets are often inadequately collected. Researchers have been advancing in the development of LLMs tailored to specific regional languages \cite{cui2023efficient} \cite{balachandran2023tamil} \cite{kunchukuttan2020ai4bharat}. For development of a robust multilingual LLM, two pivotal components are: presence of abundant multilingual data and a diverse vocabulary \cite{yuan2023multilingual}. The current state-of-art LLMs provide a rigid multilingual support, this is due to less multilingual data in the pre-training corpus. For example, Llama models \cite{touvron2023llama} \cite{touvron2023llama2} have leveraged a vast pre-training corpus with over 1.6 trillion tokens, but less than 4.5\% is multilingual data, over 20 different languages. This number is further enhanced in Llama 2 models where the proportion of multilingual data is increased to approximately 11\% and the number of languages to around 26. CulturaX \cite{nguyen2023culturax} is another multilingual dataset with 6.3 trillion tokens across 167 languages. The dataset is created after meticulously cleaning and deduplication steps to facilitate advancements in multilingual LLMs. However, out of 167 languages only 14 languages amount to 90.38\%, thus creating a very low pre-training corpus for developing an efficient and versatile LLM having contextual understanding of indic languages. 
    
    To cater this issue of procuring massive datasets for LLM development, researchers have been relying on open source datasets such as web-crawled Common Crawl, Wikipedia, Public domain books spanning from cultural and historical facets \cite{gao2020pile}, Stack Exchange and Github archives, Journal articles and educational resources \cite{lewkowycz2022solving}, News archives, Government and institutional legal repositories, Multimedia transcripts. In this study, we aim to exploit the aforementioned data sources for developing extensive indic data corpus.
    
    Moreover, these massive corpora need meticulous rigorous data refinement and deduplication to ensure data quality while maintaining integrity. Recent LLMs have demonstrated a robust work on data preparation and filtering, The RefinedWeb \cite{Penedo2023} has been pivotal in providing an insightful technical background in this context, significantly enhancing our understanding of the nuances involved. RefinedWeb has executed filtering and deduplication techniques on Common Crawl corpus, these filters include incorporation of threshold-based language filtering, url-filtering, line-wise correction filters, document deduplication, etc. MassiveText \cite{rae2021scaling} has defined rules for reducing nuances from the text documents by implementing extensive quality filtering techniques. Upon rigorous filtering and deduplication steps, a large amount of noisy data is removed from the original corpora. It has been observed that the open source language datasets do hold a high number of boilerplate texts and similar-context text documents \cite{Penedo2023} \cite{lee2021deduplicating}. 

    Consequently, deduplication process is a crucial step for producing high quality pre-training corpus. Various deduplication algorithms have been established in literature spanning from exact matching with suffix arrays \cite{manber1993suffix}, largest substring matching , minhash \cite{broder1997resemblance}, simhash \cite{charikar2002}, and other fuzzy techniques in order to minimize memory usage while maximizing efficiency.	In this work, we have methodically incorporated the data filtering and deduplication techniques, drawing inspiration from the research findings presented in RefinedWeb and MassiveText and also proposed our own innovative filters for data preprocessing.
    
    Conventional tokenization approaches often involve a complex preprocessing pipeline and are language-specific. A simplistic and multilingual tokenizer is required for diverse natural language processing (NLP) tasks. These tokenization techniques generally include BPE \cite{sennrich2015neural}, Wordpiece \cite{wu2016google} , Sentencepiece \cite{kudo2018sentencepiece}, IndicBERT \cite{kakwani2020indicnlpsuite}, Spacy \cite{spacy_tokenizer}. \cite{kunchukuttan2020ai4bharat} proposed an Indic NLP tokenizer \cite{kunchukuttan2020indicnlp} which emerges as an effective tokenization tool for indic-languages (Assamese, Bengali, Gujarati, Hindi, Marathi, Odia, Punjabi, Kannada, Malayalam, Tamil, Telugu) including English. In addition, Stanford NLP \cite{al2013speedread} has proposed Indic tokenizer \cite{manning2014stanford} which supports English, Indo-aryan and Dravidian languages along with preprocessing functionalities. SentencePiece \cite{kudo2018sentencepiece} tokenizer enables a fully end-to-end language-independent system by directly training its subword models from raw input sentences. This approach eliminates the need for pre-tokenized word sequences. In this study, we experiment with the SentencePiece \cite{kudo2018sentencepiece} tokenizer and fine-tune it based on our indic-rich corpora.
    
    In this study, we meticulously compile data corpus from open source and proprietary datasets for a comprehensive selection of 12 indic languages including English. These languages span from Assamese, Bengali, English, Kannada, Gujarati, Hindi, Marathi, Malayali, Punjabi, Odia, Tamil, Telugu. Our contributions in paper are highlighted as : \\
    1. Development of High-Quality Multilingual Indic Data \\
    2. Novel Multilingual Tokenizer Training Strategy

   % Kakde and Sawjanya

\section{Pre-training Data}
\subsection{Common Crawl}

Common Crawl is an open repository that houses extensive web crawl data. Since its inception in 2008, the archive has amassed petabytes of data and continues to perform crawls almost monthly. The data, accessible via Common Crawl's public S3 bucket, is provided in three distinct archive formats: WAT, WET, and WARC files.
The WAT (Web Archive Transformation) files encompass the metadata of the crawl, including HTTP headers, elements from the HTML <head> (such as title, meta tags, and scripts), and links from the websites. WET (Web Extracted Text) files contain the text extracted directly from the HTML of the crawl. Meanwhile, WARC (Web ARChive) files comprise the complete crawl data, encompassing both the metadata and the full HTML response.
While WET files could have served as a straightforward source for text data, our experimentation with various HTML scraping tools revealed that the text extracted in WET files often lacks cleanliness. Therefore, we opted to process the WARC files, which allowed us to scrape cleaner text from their comprehensive HTML archives.

To date, we have processed a total of 93 Common Crawl snapshots, with CC-MAIN-2023-50 being the latest snapshot at the time of writing this paper. Our processing of the Common Crawl data is divided into three major steps:
\begin{enumerate}
    \item Preprocessing : It involves raw text extraction.
    \item Postprocessing : Entails  Language Detection and the application of Heuristic Filters.
    \item Deduplication : Aimed at removing duplicate components.
\end{enumerate}

\begin{figure}[h]
\centering
\includegraphics[width=0.47\textwidth]{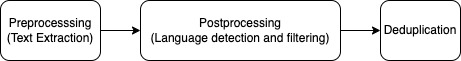}
\caption{Common Crawl Dataset Processing Pipeline}
\label{fig:cc_processing}
\end{figure}

\noindent
\textbf{Common Crawl Preprocessing}
Among the various data sources utilized during the pretraining phase, the Common Crawl dataset presented the most significant challenge, primarily due to its vast scale, which encompasses approximately 7 petabytes of data.
We employed the warcio library in Python, which is adept at efficiently streaming WARC files instead of loading them entirely into memory. This approach is crucial given the substantial size of WARC files, often reaching gigabytes, making complete loading into memory impractical.
This streaming process necessitated a separate post-processing pipeline for language identification and heuristic feature computation, which proved more efficient when conducted in batches rather than on individual records.

In our exploration of various open-source text extraction libraries, including unstructured.io and trafilatura, we found that trafilatura delivered the most effective results. While streaming through the warcio iterator, we extracted clean text, markdown text, and URLs from each web page using trafilatura.

Our analysis encompassed a total of 5,455,398 WARC files across 93 snapshots. We initially conducted a Proof of Concept (POC) for WARC preprocessing using a PySpark pipeline. In this pipeline, we read multiple WARC files as RDDs (Resilient Distributed Datasets) and processed them concurrently. Each RDD in this pipeline was a tuple containing the file path and the content of the WARC file as a byte array. We employed the same warcio library in the map partition User-Defined Function (UDF) for processing these files.

Additionally, we established a vanilla Python processing pipeline utilizing multiprocessing, which enabled parallel processing of multiple WARC files, fully utilizing the cores of a machine. The extraction of text using trafilatura from each WARC file took approximately 40 to 60 minutes, varying with the number of web pages archived in each file. Both the PySpark and multiprocessing vanilla Python pipelines demonstrated similar processing times for the WARC files. However, we opted to process the files using our multiprocessing vanilla Python pipeline. This decision was driven by the significant reduction in compute costs it offered. Furthermore, this approach allowed us to design our pipeline more fault-tolerantly, leveraging a custom orchestration pipeline that ran on multiple AWS EC2 instances.
\\

\noindent
\textbf{Common Crawl Postprocessing}
Following the preprocessing step for the WARC files, we obtain a CSV file corresponding to each WARC file. These files comprise columns detailing the scraping date, clean text, markdown text, and URL. As our focus is on developing Large Language Models (LLMs) for Indic languages, an initial step involves applying a language detection filter to segregate non-Indic languages, with the exception of English. In our exploration of various open-source language detection models, including AI4Bharat and FastText, we found that FastText provided the most accurate results. Currently, the text extracted from the Common Crawl dataset is being classified into English and 11 specific Indic languages.

Post this language filtration, approximately 50\% of the initial data remains. Subsequently, we compute heuristic features such as token count, mean sentence length, mean word length, symbol-to-word ratio, perplexity (ppl) score, fraction of duplicate lines, fraction of characters in duplicate lines, fraction of characters in the most common 2-11-grams, fraction of lines ending with an ellipsis, and fraction of lines starting with a bullet point. Through exploratory data analysis conducted for each language, we determined that applying these filters within the range of 0-90th percentiles effectively eliminates unclean and gibberish documents. Additionally, we examined the impact of applying filters solely on mean word length, mean sentence length, language threshold, and symbol-to-word ratio. The outcomes were found to be similar to those obtained when applying the full range of heuristic features.Table \ref{tab:filter-analysis} presents the number of documents and token count subsequent to the application of these filters.

\begin{table*}[h]
\centering
\captionsetup{justification=centering,margin=2cm}
\begin{tabular}{
  |>{\centering\arraybackslash}p{2cm}|
   >{\centering\arraybackslash}p{2cm}|
   >{\centering\arraybackslash}p{2cm}|
   >{\centering\arraybackslash}p{2cm}|
   >{\centering\arraybackslash}p{2cm}|
   >{\centering\arraybackslash}p{2cm}|
   >{\centering\arraybackslash}p{2cm}|
}

\hline
\textbf{Language} & \textbf{Lang(0.1) + Basic filter } & \textbf{Lang(0.6) + Basic filter }& \textbf{Dedup + Lang(0.6) + Basic filter } & \textbf{Dedup + Lang(0.6)  + All filter }\\
\hline
Hi & 108 & 85 & 54 & 51.2 \\
Bn & 49.5 & 34 & 18.6 & 16.3 \\
Ta & 33.6 & 22.3 & 6.2 & 5.1 \\
Ml & 14.3 & 9.5 & 3.2 & 2.7 \\
Mr & 11.2 & 8.1 & 2.7 & 2.4 \\
Te & 10.5 & 4 & 1.8 & 1.6 \\
Kn & 6.2 & 4.2 & 1.4 & 1.1 \\
Gu & 6.1 & 4.5 & 1.5 & 1.3 \\
Pa & 5.1 & 2.7 & 0.88 & 0.69 \\
Or & 1.2 & 0.9 & 0.4 & 0.3 \\
As & 0.8 & 0.6 & 0.01 & 0.01 \\\hline
\end{tabular}
\caption{Comparative Analysis of Token Counts(in millions) Before and After Application of Basic, Full Range Filters and Deduplication}
\label{tab:filter-analysis}
\end{table*}

\vspace{0.5cm}

\noindent
\textbf{Common Crawl Deduplication}
As noted on the Common Crawl website, each snapshot of their web crawl typically encompasses approximately 3 billion web pages. Remarkably, 2 billion of these pages have been previously crawled in earlier snapshots. This results in an average of about 66\% duplicate content per snapshot. When considering all snapshots cumulatively, the proportion of duplicate content is substantially higher. Consequently, the implementation of a deduplication process is crucial. It not only ensures the high quality of data but also significantly reduces the computational resources required for training.\\
In our deduplication pipeline, we have employed the Minhash Locality Sensitive Hashing (LSH) technique. The Minhash LSH process involves three primary steps: first is shingling, which converts documents into set representations; second is min-hashing, which transforms these large sets into shorter signatures while preserving their similarity; and finally, LSH, which identifies likely candidates for similarity based on these signatures. We have opted for 5-gram shingles for the initial conversion of each document into sets. In the min-hashing step, we set the signature dimension to 250. By establishing a similarity threshold of 0.7, we determined that the optimal configuration consists of 25 bands and 10 rows per band for this process.\\
While duplicates were present across snapshots, we chose not to run our deduplication pipeline across all snapshots to avoid excessive data loss. Additionally, we noted that if a URL is re-crawled within approximately a year, its content is likely to remain largely unchanged, although the converse may not hold true. Therefore, we segmented our deduplication process, considering snapshots spanning 1-2 years as a single batch, which roughly equates to 10 million documents. Thus, we divided the 93 snapshots for each language into batches of 10 million documents and executed the deduplication pipeline on each batch. This approach effectively removes recent duplicated data while preserving older, potentially unique data.The number of documents before and after deduplication is presented in Tabel \ref{tab:filter-analysis}

\begin{table}[h]
\centering
\captionsetup{justification=centering,}
\scriptsize
\begin{tabular}{
  |>{\centering\arraybackslash}p{1.5cm}|
   >{\centering\arraybackslash}p{1cm}|
   >{\centering\arraybackslash}p{1cm}|
   >{\centering\arraybackslash}p{1cm}|
   >{\centering\arraybackslash}p{1cm}|
   >{\centering\arraybackslash}p{1cm}|
   >{\centering\arraybackslash}p{1cm}|
}

\hline
\textbf{Language} & \textbf{Token count } & \textbf{Mean word length }& \textbf{symbol to word ratio } & \textbf{mean sentence length} \\
\hline
Hi & [50,10000] & [3 10] & 0.22 & >4.2 \\
Bn & [40,10000] & [4,9] & 0.24 & >4.4 \\
Ta & [45,10000] & [6,9] & 0.32 & >5.1 \\
Ml & [55,10000] & [6,10] & 0.25 & >3.5 \\
Mr & [43,10000] & [4,6] & 0.31 & >4.3 \\
Te & [52,10000] & [5,8] & 0.28 & >5.4 \\
Kn & [48,10000] & [5,8] & 0.32 & >3.6 \\
Gu & [51,10000] & [4,6] & 0.23 & >3.4 \\
Pa & [55,10000] & [3,6] & 0.24 & >3.7 \\
Or & [47,10000] & [4,7] & 0.32 & >4.2 \\
As & [49,10000] & [4,7] & 0.25 & >3.8 \\\hline
\end{tabular}
\caption{Comparative Analysis of Filter Thresholds Language-wise}
\label{tab:dedup-stats}
\end{table}  % Rahul and Divyansh
\subsection{Newspaper}

There are plethora of newspapers across various indic languages which are regarded as a knowledge-rich source of data for pre-training our indic-rich context LLM. We accumulated language-specific newspapers and downloaded the digital version of all the historical editions. These newspapers generally comprise blocks of images, tables and advertisements which are noisy data in pre-training corpora. Henceforth, a robust and scalable algorithm is necessary to identify individual useful text-blocks from the news article and subsequently extract data out of it. This elementary problem of article extraction is challenging due to composition of a wide range of layouts random arranged in the target page. Research has been prevailing for development of algorithms addressing the text extraction issues . However, most of these methods are based on a set of heuristic rules. 
\noindent
In this study, we construct an article-extraction pipeline by experimenting with existing open source algorithms and frameworks. Our goal in this extraction process is to identify bounding boxes of layouts in an article and then detect relevant text-blocks for respective languages. Initially, we experiment with an open source python package, layoutparser, prolific in document-image analysis tasks. LayoutParser provides a rich repository of deep learning models for layout detection as well as a set of unified APIs for using them.LayoutParser comes with a set of layout data structures with carefully designed APIs that are optimized for document image analysis tasks. In addtion, we train models on indic-data comprising of various handwritten documents and newspapers. In this step, we experiment with deep learning-based models by detectron2 from layoutparser library are leveraged to detect text-snippets from the layouts of the article. Furthermore, this algorithm also detects tables which are redundant for the text corpus. After rigorous experimentation, we resolved to fine-tune a MaskRCNN based model for our extraction process.
% The overall architecture for the model fine-tuning and extraction is depicted in Fig. :

% \includegraphics[width=0.4\textwidth]{architecture.eps}

\vspace{0.5cm}
\noindent
\textbf{Annotation}
All the segmented blocks of the newspaper images are labelled into 4 classes namely headings, text, images and non-text by leveraging label studio. The text class focuses mainly on the article block and this is what we intend to extract. The headings class contains all the main and sub headings in the newspaper image. The non-text class taks care of the tables, summaries and quotes. %The annotation of a sample image is depicted in Fig.:

\vspace{0.5cm}
\noindent
\textbf{Model fine-tuning}
MaskRCNN based model is a three stage model that predicts a class label, bounding box offset and object mask for each Region of Interest (ROI). It can produce more accurate and fine-grained masks for each object, which can capture the shape and contour details better than bounding boxes. It can also handle overlapping and occluded objects better. %The architecture of MaskRCNN is depicted in Fig. 

% The details regarding the final training and evaluation data is depicted in the table:
% The training loss curve is depicted in fig. 

\vspace{0.5cm}
\noindent
\textbf{Model evaluation}
The performance of the object detection and localization algorithm is evaluated by a metric called Average Precision (AP) and mean average precision (MAP).
AP is calculated with the help of several other metrics such as Intersection over Union (IoU), confusion matrix (TP, FP, FN), precision and recall as shown in the figure below.
IoU quantifies the closeness of the two bounding boxes (ground truth and prediction). It's a value between 0 and 1.
The IoU is calculated by taking the ratio between the area of intersection and the area of the union of two bounding boxes.
Average Precision is the area under the PR curve. AP summarizes the PR Curve to one scalar value. Average precision is high when both precision and recall are high, and low when either of them is low across a range of confidence threshold values. The range for AP is between 0 to 1.
AP value can be calculated for each class. The mean average precision is calculated by taking the average of AP across all the classes under consideration.
%The model's evaluation across all the classes is depicted below.

\vspace{0.5cm}
\noindent
\textbf{Model Inference}
The model inference pipeline takes up one newspaper image and then bounding box prediction is done on the entire newspaper image along with the class. Masking of the heading, non-text and images class bounding boxes is done on the image. For each text class bounding box, a tesseract based OCR is used for extraction of text. %This is depicted in fig. 

  % Rishabh
\subsection{Books}

The web crawled data from Common Crawl has limited distribution of Indic language specific content. The distribution of deduplicated content is indeed further lesser. This is a huge challenge especially when we want to perform LLM pre-training with Indic languages. In order to increase the number of tokens as well as the breadth of Indic specific content, we leverage the open source pdfs that mostly contain books, periodicals, magazines, and financial reports. The pdfs are downloaded using a pipeline across the indic languages.\\
\noindent
The pdfs have broadly two types of format present in it, image or text embedded on pdf. A pipeline is built that detects whether an image is embedded on the pdf and then based on that, appropriate extraction pipeline runs. Tesseract OCR is leveraged to extract text out of these pdf documents with images embedded on it. There is a challenge of detecting a script so that tesseract OCR engine with appropriate language is used for extraction. We overcome this with our pipeline that detects the script and extracts text out of it. It is then scaled up with multiprocessing and across multiple nodes.  % Rishabh and Kakde

\section{Tokenizer}

Tokenization is a preprocessing step in natural language processing (NLP), facilitating machine comprehension and interpretation of human language by breaking down text into fundamental units such as words, characters and subwords. Among the various types of tokenization tools, one widely employed approach is SentencePiece, which sets itself apart by directly training subword models from raw sentence data. This method offers a fully end-to-end and language-agnostic system, eliminating the need for pre-tokenized word sequences. 

In our study, we adopted Byte Pair Encoding (BPE) as the tokenization strategy. We assessed the quality of tokenization using two key metrics: the token-to-word ratio and the exact score, which measures the proportion of correctly tokenized units out of the total tokens. 

\begin{table}[ht]
    \centering
    \begin{tabular}{|c|c|c|c|}
        \hline
        Language & 70k & 85k & 100k \\ \hline
        en & 1.52 & 1.49 & 1.46 \\ \hline
        hi & 1.38 & 1.36 & 1.35 \\ \hline
        mr & 1.83 & 1.83 & 1.79 \\ \hline
        te & 2.58 & 2.45 & 2.39 \\ \hline
        ta & 2.51 & 2.41 & 2.30 \\ \hline
        as & 2.06 & 2.03 & 2.00 \\ \hline
        gu & 2.05 & 1.92 & 1.90 \\ \hline
        kn & 2.51 & 2.32 & 2.29 \\ \hline
        ml & 3.01 & 2.98 & 2.86 \\ \hline
        od & 1.94 & 1.89 & 1.85 \\ \hline
        pn & 1.58 & 1.66 & 1.63 \\ \hline
    \end{tabular}
     \caption{Token-to-Word Ratio for Different Vocabulary Sizes}
    \label{tab:token-to-word-ratio}
\end{table}

\subsection{Tokenizer Experiments}
\noindent
\textbf{Corpus Size}
For training the tokenizer, we utilized sample data from various sources such as Common Crawl, Wikipedia, and Books. This sampled data constitutes the corpus size employed during tokenizer training. The data in the corpus is sampled in proportion to the availability of data from these different sources.
\noindent
The vocabulary size in the context of a tokenizer refers to the number of unique tokens or subword units that the tokenizer can produce or handle. We trained the tokenizer with different corpus sizes and vocabulary sizes. The results in \textbf{corpus and vocab table} indicate that increasing the corpus size does not significantly improve the performance of the tokenizer. However, it is evident that increasing the vocabulary size significantly improves the token-to-word ratio.
\noindent
This finding is logical because increasing the corpus size does not necessarily enhance tokenizer performance once a critical mass of major words has been covered. If the vocabulary size remains constant, the same or similar tokens will be generated, and the token-to-word ratio will remain unchanged, even with an increased corpus size. Thus, beyond a certain point, expanding the corpus size yields diminishing returns in terms of tokenizer improvement.

\vspace{0.5cm}
\noindent
\textbf{Vocabulary Size}
Additionally, we experimented with various corpus sizes. The results, as presented in the \textbf{corpus table}, demonstrate that a tokenizer trained with a 225 million sampled corpus size achieves a token-to-word ratio similar to that of a tokenizer trained with a 12 billion sampled corpus size.
Increasing the vocabulary size allows for the formation of more eligible words or sub-words as tokens, thereby improving the tokenizer's quality, as indicated by the improved token-to-word ratio. However, this improvement comes with a trade-off: a tokenizer with a larger vocabulary size will have reduced inference speed compared to one with a smaller vocabulary size.

\vspace{0.5cm}
\noindent
\textbf{Character Coverage}
We also experimented with different character coverage levels during tokenizer training. Setting character coverage to 1 resulted in the inclusion of numerous gibberish characters, such as emojis and special characters, in the tokenizer vocabulary. These characters, despite having low frequencies, were included due to the high character coverage. To address this, we tested various character coverage settings. As presented in the \textbf{character coverage table}, we found that a character coverage of 0.997 is optimal. This setting ensures that the characters included in the Hindi tokenizer vocabulary closely match the typical set of characters in the Hindi language, which consists of approximately 50 characters (36 consonants and 12 vowels). 

\begin{table}[ht]
    \centering
    \begin{tabular}{|c|c|c|}
        \hline
        Language & GPT-4o & Indic \\ \hline
        en & 1.33 & 1.48 \\ \hline
        hi & 1.62 & 1.39 \\ \hline
        mr & 2.53 & 1.80 \\ \hline
        te & 3.15 & 2.29 \\ \hline
        ta & 3.16 & 2.30 \\ \hline
        as & 2.70 & 1.99 \\ \hline
        gu & 2.28 & 1.91 \\ \hline
        kn & 3.03 & 2.31 \\ \hline
        ml & 3.37 & 2.71 \\ \hline
        od & 6.39 & 1.80 \\ \hline
        pn & 2.68 & 1.56 \\ \hline
    \end{tabular}
      \caption{Token-to-Word Ratios for GPT-4o and Indic 100k tokenizer on ai4bharat sangraha Corpus}
    \label{tab:Tokenizer comparision}
\end{table}

\subsection{Tokenizer Training}

Initially, we trained a tokenizer with the optimal vocabulary size, corpus size, and character coverage determined from our previous experiments. Despite sampling a dataset focused on Indic content, we observed a significant number of false positives, which introduced characters from other languages, such as Chinese and French, into our Indic LLMs. To address this issue and create a clean Indic tokenizer, we implemented a series of steps specifically targeting Indic words.

First, we trained a dummy tokenizer using the sampled data from various sources. With the assistance of linguistic experts, we manually removed gibberish characters and characters from other languages from the dummy tokenizer vocabulary. Next, we encoded and decoded the sampled dataset using this dummy tokenizer, with the byte fallback flag set to false. This process converted all gibberish words and non-Indic language words identified by linguists to UNK tokens.

With this cleaned dataset, we then trained our final tokenizer. This refined tokenizer demonstrated optimal vocabulary size, corpus size, and character coverage, with minimal inclusion of gibberish and foreign language words. We compared the performance of our tokenizer with the OpenAI Tiktoken tokenizer across 11 Indic languages. The results, as shown in the Table 4, clearly indicate that our tokenizer's metrics significantly outperform those of the Tiktoken tokenizer.

\if dfd
\begin{table*}[h]
\centering
{%
\begin{tabular}{|>{\centering\arraybackslash}p{2cm}|>{\centering\arraybackslash}p{0.8cm}|>{\centering\arraybackslash}p{0.8cm}|>{\centering\arraybackslash}p{0.8cm}|>{\centering\arraybackslash}p{0.8cm}|>{\centering\arraybackslash}p{0.8cm}|>{\centering\arraybackslash}p{0.8cm}|>{\centering\arraybackslash}p{0.8cm}|>{\centering\arraybackslash}p{0.8cm}|>{\centering\arraybackslash}p{0.8cm}|>{\centering\arraybackslash}p{0.8cm}|>{\centering\arraybackslash}p{0.8cm}|}

\hline
Vocab Size&
en&
hi&
mr&
te&
ta&
as&
gu&
kn&
ml&
od&
pn
\\
\hline
70k& 
1.52&
1.38&
1.83&
2.58&
2.51&
2.06&
2.05&
2.51&
3.01&
1.94&
1.58
\\
\hline
85k& 
1.49&
1.36&
1.83&
2.45&
2.41&
2.03&
1.92&
2.32&
2.98&
1.89&
1.66
\\
\hline
100k& 
1.46&
1.35&
1.79&
2.39&
2.30&
2.00&
1.90&
2.29&
2.86&
1.85&
1.63
\\
\hline
\end{tabular}
}
\caption{Token-to-Word Ratios for Different Tokenizer Vocabulary Sizes on Wiki Corpus}

\end{table*}
\FloatBarrier

\begin{table*}[h]
\centering
{%
\begin{tabular}{|>{\centering\arraybackslash}p{2cm}|>{\centering\arraybackslash}p{0.8cm}|>{\centering\arraybackslash}p{0.8cm}|>{\centering\arraybackslash}p{0.8cm}|>{\centering\arraybackslash}p{0.8cm}|>{\centering\arraybackslash}p{0.8cm}|>{\centering\arraybackslash}p{0.8cm}|>{\centering\arraybackslash}p{0.8cm}|>{\centering\arraybackslash}p{0.8cm}|>{\centering\arraybackslash}p{0.8cm}|>{\centering\arraybackslash}p{0.8cm}|>{\centering\arraybackslash}p{0.8cm}|}

\hline
Tokenizer &
en &
hi &
mr &
te &
ta &
as &
gu &
kn &
ml &
od &
pn \\
\hline
GPT-4o & 
1.33 &
1.62 &
2.53 &
3.15 &
3.16 &
2.70 &
2.28 &
3.03 &
3.37 &
6.39 &
2.68 \\
\hline
Indic & 
1.48 &
1.39 &
1.80 &
2.29 &
2.30 &
1.99 &
1.91 &
2.31 &
2.71 &
1.80 &
1.56 \\
\hline
\end{tabular}
}
\caption{Token-to-Word Ratios for GPT-4o and Indic 100k tokenizer on ai4bharat sangraha Corpus}
\end{table*}
\FloatBarrier

\fi

 % Sowjanya
% \input{files/tokenizer_old.tex} % Sowjanya

% \input{files/experiments_and_results.tex}  % Daud Ibrahim

\section{Conclusion}

In conclusion, our meticulous data acquisition and custom preprocessing pipeline have effectively curated a high-quality multilingual dataset, significantly enhancing the performance of our Indic large language models. Through rigorous filtering and deduplication processes, we ensured the elimination of redundant and low-quality content, optimizing the data for training. Our novel multilingual tokenizer training strategy demonstrated superior token-to-word ratios for Indic languages compared to the state-of-the-art OpenAI Tiktoken tokenizer. The experiments underscore the importance of tailored preprocessing and tokenizer design, paving the way for more accurate and efficient language models in Indic-rich multilingual contexts.  % Rahul

% \section*{Acknowledgements}

\bibliography{custom}

\end{document}